\documentclass{article} 
\usepackage{iclr2025_conference,times}

\usepackage{amsmath,amsfonts,bm}

\def\eqref#1{equation~\ref{#1}}

\def\1{\bm{1}}

\DeclareMathAlphabet{\mathsfit}{\encodingdefault}{\sfdefault}{m}{sl}
\SetMathAlphabet{\mathsfit}{bold}{\encodingdefault}{\sfdefault}{bx}{n}

\usepackage{hyperref}
\usepackage{url}

\usepackage{booktabs}
\usepackage{pifont}
\usepackage{multirow}

\usepackage{graphicx}
\usepackage{booktabs}
\usepackage{pifont} 

\newcommand{\cmark}{\ding{51}} 
\newcommand{\xmark}{\ding{55}} 

\usepackage{graphicx}
\usepackage{xcolor}

\definecolor{lightblue}{rgb}{0.68, 0.85, 0.9}  
\definecolor{lightgreen}{rgb}{0.56, 0.93, 0.56}  
\usepackage{subcaption}
\usepackage{wrapfig}
\usepackage{float}
\usepackage{microtype}
\usepackage{graphicx}
\usepackage{booktabs} 
\usepackage{hyperref}
\usepackage{multirow}
\usepackage{makecell}
\usepackage{amsmath}
\usepackage{amssymb}
\usepackage{mathtools}
\usepackage{amsthm}
\usepackage{url}            
\usepackage{amsfonts}       
\usepackage{nicefrac}       
\usepackage{microtype}      
\usepackage{xcolor}    
\usepackage{multirow}
\usepackage{array}
\usepackage{ifthen}
\usepackage{bbm}
\newboolean{include-notes}
\usepackage{amsfonts}

\usepackage{amsthm}
\newtheorem{definition}{Definition}
\usepackage{wrapfig}
\usepackage{caption}
\usepackage{tikz}
\usepackage{authblk}
\usepackage{hyperref}
\hypersetup{colorlinks=true, linkcolor=blue, citecolor=black} 
\title{Evaluating LLMs Without Oracle Feedback: Agentic Annotation Evaluation Through Unsupervised Consistency Signals}
\author[1,3]{Cheng Chen} \author[3]{Haiyan Yin} \author[2,3]{Ivor W. Tsang}
\affil[1]{ Australian Artificial Intelligence Institute (AAII), University of Technology Sydney, Australia}
\affil[2]{ College of Computing and Data Science, Nanyang Technological University, Singapore}
\affil[3]{ Centre for Frontier AI Research, Institute of High Performance Computing\\ 
Agency for Science, Technology and Research, Singapore}
\affil[ ]{ \texttt{Cheng.chen-16@student.uts.edu.au,}  \texttt{\{ivor$\_$tsang, yin$\_$haiyan\}@cfar.a-star.edu.sg}}

\iclrfinalcopy 
\begin{document}
\maketitle
\vspace{-0.25in}
\begin{abstract}
Large Language Models (LLMs), when paired with prompt-based tasks, have significantly reduced data annotation costs and reliance on human annotators. However, evaluating the quality of their annotations remains challenging in dynamic, unsupervised environments where oracle feedback is scarce and conventional methods fail.
To address this challenge, we propose a novel agentic annotation paradigm, where a student model collaborates with a noisy teacher (the LLM) to assess and refine annotation quality without relying on oracle feedback. The student model, acting as an unsupervised feedback mechanism, employs a user preference-based majority voting strategy to evaluate the consistency of the LLM’s outputs. 
To systematically measure the reliability of LLM-generated annotations, we introduce the \textbf{Consistent and Inconsistent (CAI)} Ratio, a novel unsupervised evaluation metric. The CAI Ratio not only quantifies the annotation quality of the noisy teacher under limited user preferences but also plays a critical role in model selection, enabling the identification of robust LLMs in dynamic, unsupervised environments.
Applied to ten open-domain NLP datasets across four LLMs, the CAI Ratio demonstrates a strong positive correlation with  LLM accuracy, establishing it as an essential tool for unsupervised evaluation and model selection in real-world settings.
\end{abstract}
\vspace{-0.3in}
\section{Introduction}
Large Language Models (LLMs), when combined with prompt optimisation \citep{brown2020language, chen2024selfteaching, kojima2022large,wei2021finetuned, wei2022chain, huang2022large, yao2022react, diao2023active, liu2023graphprompt, wang2023h, yao2024tree, long2023large, huang2023large, madaan2024self, shinn2024reflexion}, have demonstrated remarkable capabilities in text and data annotation across diverse open-domain tasks \citep{meng2022generating,ye2022zerogen,wang2024codeclm,liu2024best,wu2024unigen}, including spoken language understanding \citep{chen2024low}. Often outperforming traditional crowdsourcing and manual annotation methods~\citep{gilardi2023chatgpt}, LLM-generated annotations have become pivotal for supervised fine-tuning, alignment training, and real-time inference \citep{tan2024large}. 
However, evaluating the quality of LLM-generated annotations remains challenging in unsupervised environments where oracle feedback is unavailable. Traditional evaluation methods (Table \ref{tab:metrics_comparison} in Appendix) fall short in these settings, and LLMs often exhibit overconfidence and inconsistent behavior without external supervision \citep{xiong2023can, zhou2024relying}, underscoring  the need for robust unsupervised evaluation strategies.

\begin{figure}[t!]
\vspace{-25pt} 
    \centering
    \begin{minipage}[t]{0.25\linewidth} 
        \centering
        \includegraphics[width=\linewidth]{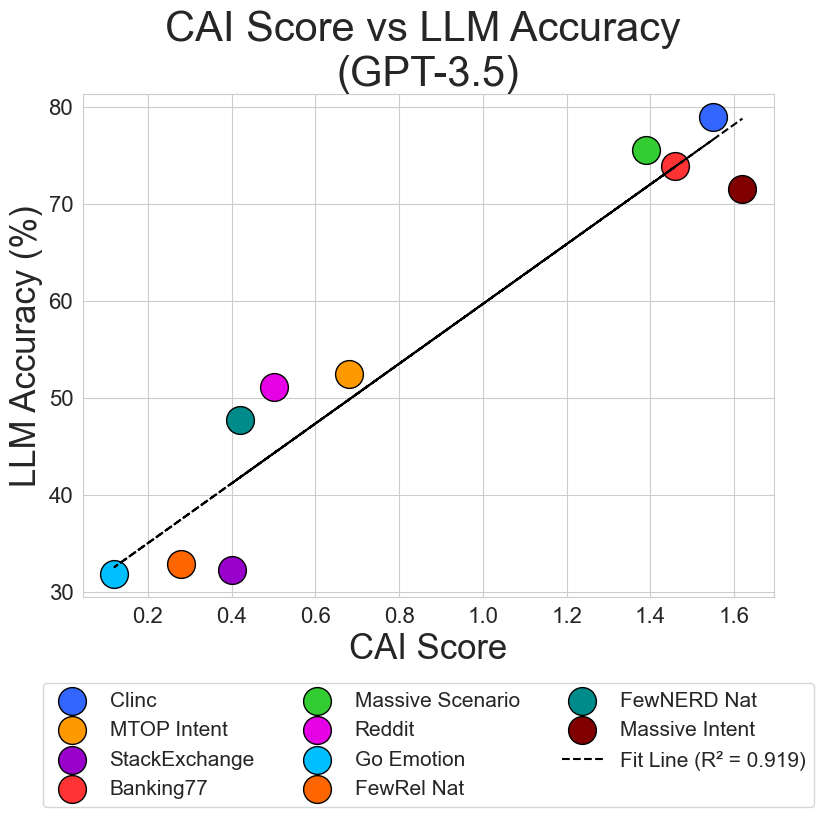}
    \end{minipage}%
    \hfill
    \begin{minipage}[t]{0.25\linewidth} 
        \centering
        \includegraphics[width=\linewidth]{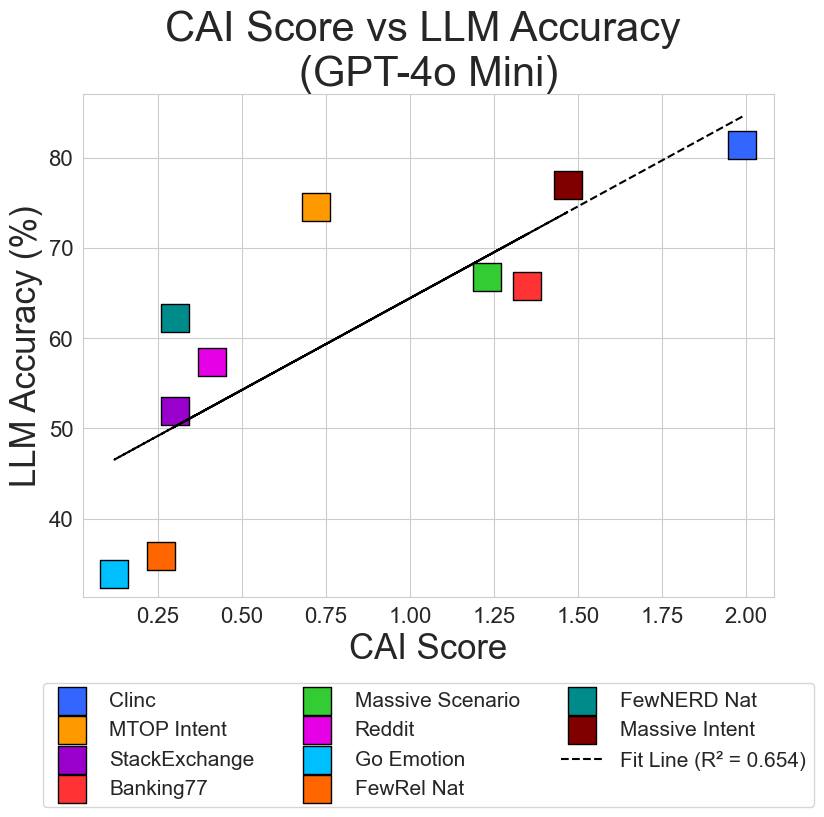}
    \end{minipage}%
    \hfill
    \begin{minipage}[t]{0.25\linewidth} 
        \centering
        \includegraphics[width=\linewidth]{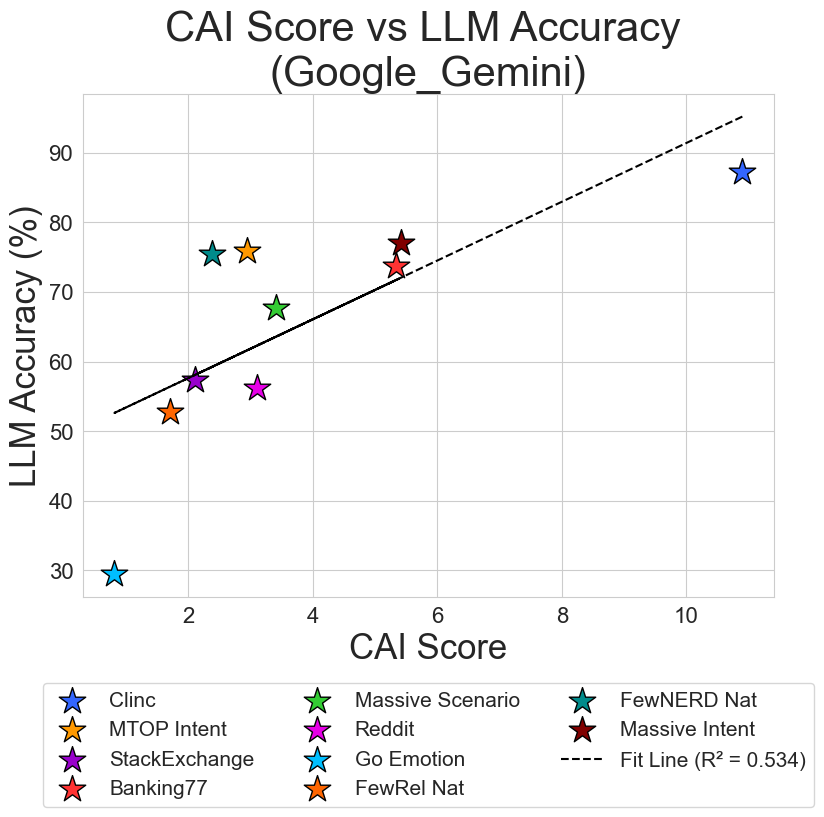}
    \end{minipage}%
    \hfill
    \begin{minipage}[t]{0.25\linewidth} 
        \centering
        \includegraphics[width=\linewidth]{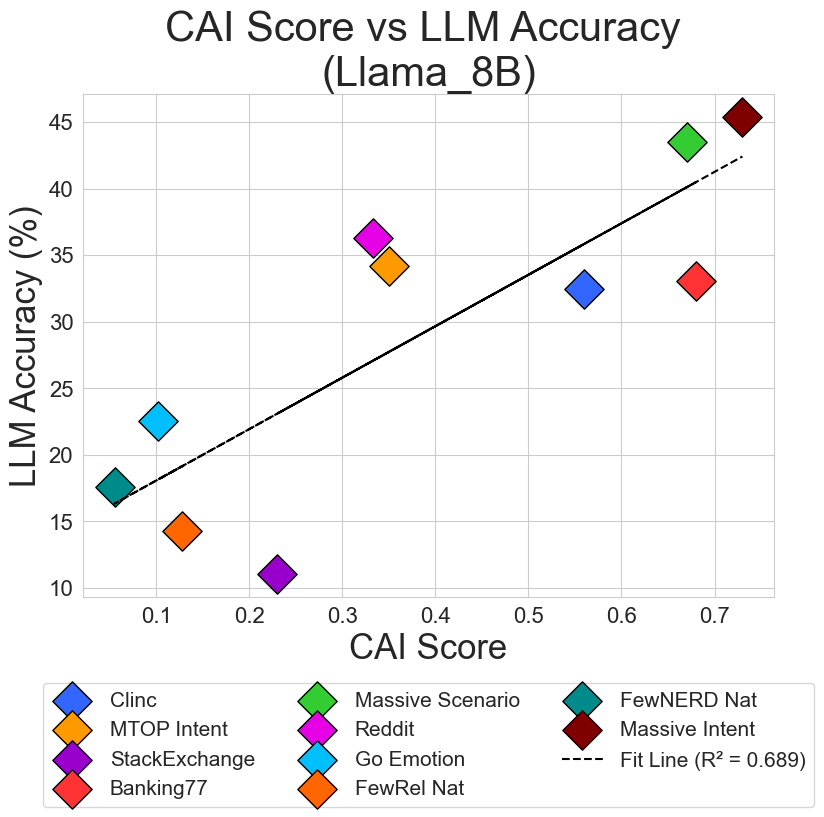}
    \end{minipage}
    \vspace{-0.2in} 
    \caption{\small\textbf{Correlation analysis between LLM annotation accuracy and the CAI ratio}, evaluated across 4 principled LLMs (also see statistical test results in Sec~\ref{sec:exp_CAI_ratio}). The Pearson correlation coefficients and corresponding p-values confirm the statistical significance of the positive correlation between CAI ratio and LLMs accuracy. }
    \label{fig:cai_correlation}
\end{figure}

To address this, we propose a novel agentic annotation evaluation paradigm, where a student model collaborates with a noisy teacher (the LLM) to assess annotation quality through model agreement. This paradigm embodies the essence of agentic reasoning: in the absence of oracle feedback, \textit{reliability emerges from the consistency of interactions between models,} with agreement serving as an implicit signal of annotation quality. 
When external supervision is missing, the alignment or misalignment between the student and the LLM offers a self-regulating mechanism to gauge annotation reliability. Building on this, we introduce the \textbf{Consistent and Inconsistent (CAI)} Ratio, a novel metric that effectively evaluates the LLM reliability by exploiting the unsupervised structural patterns within the data through the student model. By harnessing the intrinsic consistency patterns between both models, the CAI Ratio provides a powerful unsupervised signal for assessing LLM annotations. Beyond evaluation, it also serves as decisive criterion for model selection, enabling the identification of most appropriate LLMs without relying on oracle feedback. We demonstrate in Figure~\ref{fig:cai_correlation} and Table~\ref{tab:model_selection} that the CAI Ratio exhibits a strong positive correlation with LLM annotation accuracy and effectively  distinguishes the best-performing LLM models in unsupervised settings.
\vspace{-0.1in}
\section{Methodology}
We propose an agentic annotation evaluation paradigm to assess LLM reliability through the collaboration between a noisy LLM teacher and a student model. To enable annotation in an unsupervised setting, we leverage the student model’s ability to capture the structural relationships in data, assigning annotations through a majority voting mechanism in its embedding space (Equation~\ref{AS}\&~\ref{equ:majority}). Meanwhile, the LLM generates outputs via an autoregressive process. Consistent samples emerge when both models agree, indicating reliable annotations, while inconsistent samples reflect the LLM’s overconfident outputs that diverge from the student’s predictions. By systematically analyzing these agreement patterns, we capture both confidence signals and overconfidence biases, enabling robust unsupervised evaluation through the proposed Consistent and Inconsistent (CAI) Ratio.

\subsection{Problem Definition}
Given unsupervised text corpus distributions for training and testing, denoted as  $\mathcal{D}_{U} = \{x_{1}, \dots, x_{N}\}, \mathcal{D}_{U_t} = \{x'_{1}, \dots, x'_{L}\}, x, x' \in \mathcal{X} \subseteq \mathbb{R}^{d}.$ Additionally, since annotations are generated according to user preferences, a small-size user-preference distribution is provided, denoted as:  $H = \{(x_{i}, \Bar{y}_{i})\}_{i=1}^{s}, s = 5\% \times |\mathcal{D}_{U}|.$ These samples are clustered into $ k $ non-overlapping clusters $ C_1, C_2, ..., C_k $. A set of preference annotations is denoted as $ \mathcal{A}=\{\Bar{y}_{1},\Bar{y}_{2},...,\Bar{y}_{k}\}.$ Each cluster $ C_j $ is defined as  
$C_{j} = \{ x_{i} \mid x_{i} \in H_{j} \},   H_j \subseteq H,   \forall x_{i} \in C_j,   \Bar{y}_{i} = \Bar{y}_{j}.$ The clusters are disjoint, satisfying  $\bigcup_{j=1}^{k} C_j = H,   C_i \cap C_j = \emptyset,   \forall i \neq j.$ The goal is to evaluate quality of LLMs assigned annotation by estimating the latent consistent and inconsistent sets $ \mathcal{C}^* $ and $ \mathcal{I}^* $.
\subsection{Agentic  Annotation Evaluation  Through a noisy LLM Teacher and Student Model Collaboration}
\begin{wrapfigure}{r}{0.4\textwidth} 
    \centering
    \vspace{-22pt} 
    \includegraphics[width=1.05\linewidth]{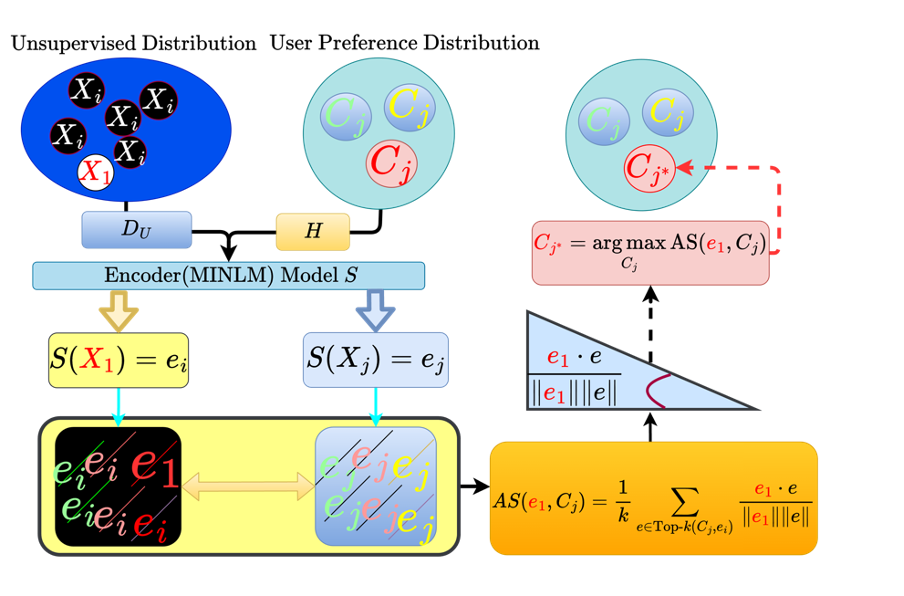}
    \vspace{-0.2in} 
    \caption{\small \textbf{Agentic Annotation Evaluation Process}: the teacher LLM generates noisy annotations in zero-/single-shot settings, which are compared against student model’s \textbf{AS} output to measure the consistency score.}
    \label{fig:cai_ratio}
    \vspace{-0.2in} 
\end{wrapfigure}
An \textbf{agentic annotation evaluation process} is constructed upon an LLM teacher with noisy annotation labels and a student model with limited user preferences, incorporating a \textit{preference-based majority voting} mechanism. We utilize MINILM \citep{wang2020minilm} as the student model, denoted as $\mathcal{S}$, which is a sentence-transformer designed for efficient sentence encoding. The student model encodes each input instance $x_i$ into its corresponding \textit{sentence embedding} as $\mathcal{S}(x_i) = e_i.$
To assign annotations, the student model employs a \textit{user preference-based majority voting} strategy, leveraging our proposed \textbf{Average Similarity (AS)} function, which is defined as:
\begin{align}
 {
    AS(e_i, C_j) = \frac{1}{k} \sum_{e \in \text{Top-}k(C_j, e_i)} \frac{e_i \cdot e}{\|e_i\| \|e\|},}
\label{AS}
\end{align}
{where $ e_{i} $ denotes the embedding for $ x_{i} $, and $ e $ represents the embedding of each sample in cluster $ C_{j} $. The term $\text{Top-}k(C_j, e_i)$ refers to the subset of samples in $C_{j}$ with the top $ k $ cosine similarity scores with $ e_{i} $. 
Formally, $\text{Top-}k(C_j, e_i) = \{e \in C_j \mid \text{AS}(e_i, e) \text{ ranks among the top } k \text{ in } C_j\}.$ Based on the computed similarity scores, the  most similar examples to $ e_{i} $ are identified, and the average cosine similarity is computed for the top-selected samples in each cluster. In our experiments, we set $ k $ to five.}
Lastly, for the annotation assignment, we assign the annotation of the cluster $C_{j}$ with the highest average cosine similarity score to the unlabelled sample $x_{i}\in D_u $.
The cluster $ C_{j^{*}} $, which has the highest average cosine similarity with the embedding $ e_i $ of a sample $ x_i $, is defined as:
\begin{align}
C_{j^{*}} = \underset{C_j}{\arg\max} \, \text{AS}(e_i, C_j),
\label{equ:majority}
\end{align}
where $\text{AS}(e_i, C_j) $ is the average cosine similarity of $ e_i $ with the embeddings in $ C_j $. The annotation $ \Bar{y}_{j^{*}} $ associated with $ C_{j^{*}} $ is then assigned to $ x_i $, i.e., $ \Bar{y}_i = \Bar{y}_{j^{*}} $. This process is represented by the annotation assignment function $h(x_i)$. Subsequently, the annotation associated with $ C_{j^{*}} $ is as defined by the user, is assigned to $ x_i $. Finally, the student agent-annotated dataset is constructed as $ D_s = \{(x_{i}, \Bar{y}_{i})\}_{i}^{N} $, where each $ \Bar{y}_{i} $ represents the student annotation obtained using the user preference-based majority voting approach. 
Given the acquired dataset $ D_{s} = \{(x_{i}, \Bar{y}_{i})\}_{i=1}^{N}$ generated by the SA, we further leverage a noisy teacher LLM to generate annotations through a \textit{group prompting} mechanism, applying both \textit{zero-shot} and \textit{single-shot} strategies. Specifically, in the zero-shot setting, 
the noisy teacher LLM generates annotations independently, defined as $ \Bar{y}^{t}_{i} = T(x_{i}) $. $\bar{y}^t_i = T(x_i)   \text{with}   P\bigl(\bar{y}^t_i \mid x_i\bigr) = \prod_{t=1}^{T_i} P\bigl(\bar{y}^t_{i,t} \mid x_i, \bar{y}^t_{i,1}, \ldots, \bar{y}^t_{i,t-1}\bigr).$ In contrast, the \textit{single-shot} setting incorporates student-generated annotations as additional context, yielding $ \hat{y}^{t}_{i} = T(x_{i}, \Bar{y}_{i}) $, where $ (x_{i}, \Bar{y}_{i}) \in D_{s} $. $\hat{y}^t_i = T(x_i, \bar{y}_i)   \text{with}   P\bigl(\hat{y}^t_i \mid x_i, \bar{y}_i\bigr) = \prod_{t=1}^{T_i} P\bigl(\hat{y}^t_{i,t} \mid x_i, \bar{y}_i, \hat{y}^t_{i,1}, \ldots, \hat{y}^t_{i,t-1}\bigr).$
Since the LLM follows an autoregressive generation framework, we query the noisy teacher LLM to provide the annotation for each instance $ x_i$ without including the student labels $ \Bar{y}_{i} $ for zero-shot prompting, producing the noisy teacher distribution $ D_{t} = \{(x_{i}, \Bar{y}^{t}_{i})\}_{i=1}^{N} $. For the single-shot setting, the SA annotations are incorporated, resulting in the student-noisy teacher distribution $ \hat{D}_{t} = \{(x_{i}, \hat{y}^{t}_{i})\}_{i=1}^{N}$. 
\vspace{-0.15in}
\subsection{Evaluation of LLMs without Oracle Feedback} 
\vspace{-0.1in}
\label{sec:evaluation}
After acquiring the SA-generated dataset \( D_{s} \), the Noisy Teacher-generated dataset \( D_{t} \), and the SA-Noisy Teacher dataset \( \hat{D}_{t} \), we introduce the Consistent-and-Inconsistent (CAI) Identification and Ratio framework. Specifically, CAI Identification determines consistent and inconsistent samples across \( D_{s} \), \( D_{t} \), and \( \hat{D}_{t} \) by comparing annotation agreement between the Student Agent (SA) \( \mathcal{S} \) and the Noisy Teacher (NT) \( \mathcal{T} \). Samples with identical predictions from both the SA and NT models are classified as consistent samples; otherwise, they are considered inconsistent samples. For each \( x \in \mathcal{D}_{u} \), the annotation assignment process is represented by the function \( h \). The annotation label from the SA is denoted as \( \bar{y}_{\mathcal{S}} \), while the NT’s annotation labels are represented as \( \bar{y}_{\mathcal{T}} \) (zero-shot) and \( \hat{y}_{\mathcal{T}} \) (single-shot). A sample is classified as consistent if $\bar{y}_{\mathcal{S}} = \bar{y}_{\mathcal{T}} = \hat{y}_{\mathcal{T}},  x \in \mathcal{C}$ where \( \mathcal{C} \) denotes the set of consistent samples. Conversely, a sample is classified as inconsistent if at least one of the assigned annotations differs, represented as $\exists \, (y, y') \in \{\bar{y}_{\mathcal{S}}, \bar{y}_{\mathcal{T}}, \hat{y}_{\mathcal{T}} \},  y \neq y',  x \in \mathcal{I}$,
where $\mathcal{I}$ represents the set of inconsistent samples. Identifying annotation inconsistencies is crucial, as is rigorously assessing the teacher model’s annotation quality, especially in the absence of ground truth. 
\begin{definition}[Consistent-and-Inconsistent (CAI) Ratio]
Let $ N_C $ and $ N_{IC} $ denote the number of \textit{consistent samples} (LLM and student model agree)
and the number of \textit{inconsistent samples} (LLM and student model disagree), 
respectively. The \text{CAI Ratio} is defined as $\text{CAI Ratio} = \frac{N_C}{N_{IC}}$.
\end{definition}
\vspace{-5pt} 
\begin{figure}[t]
\vspace{-20pt} 
\tiny
    \begin{minipage}[c]{0.50\textwidth}
        \centering
        \vskip 0.35in
        \includegraphics[width=0.89\linewidth]{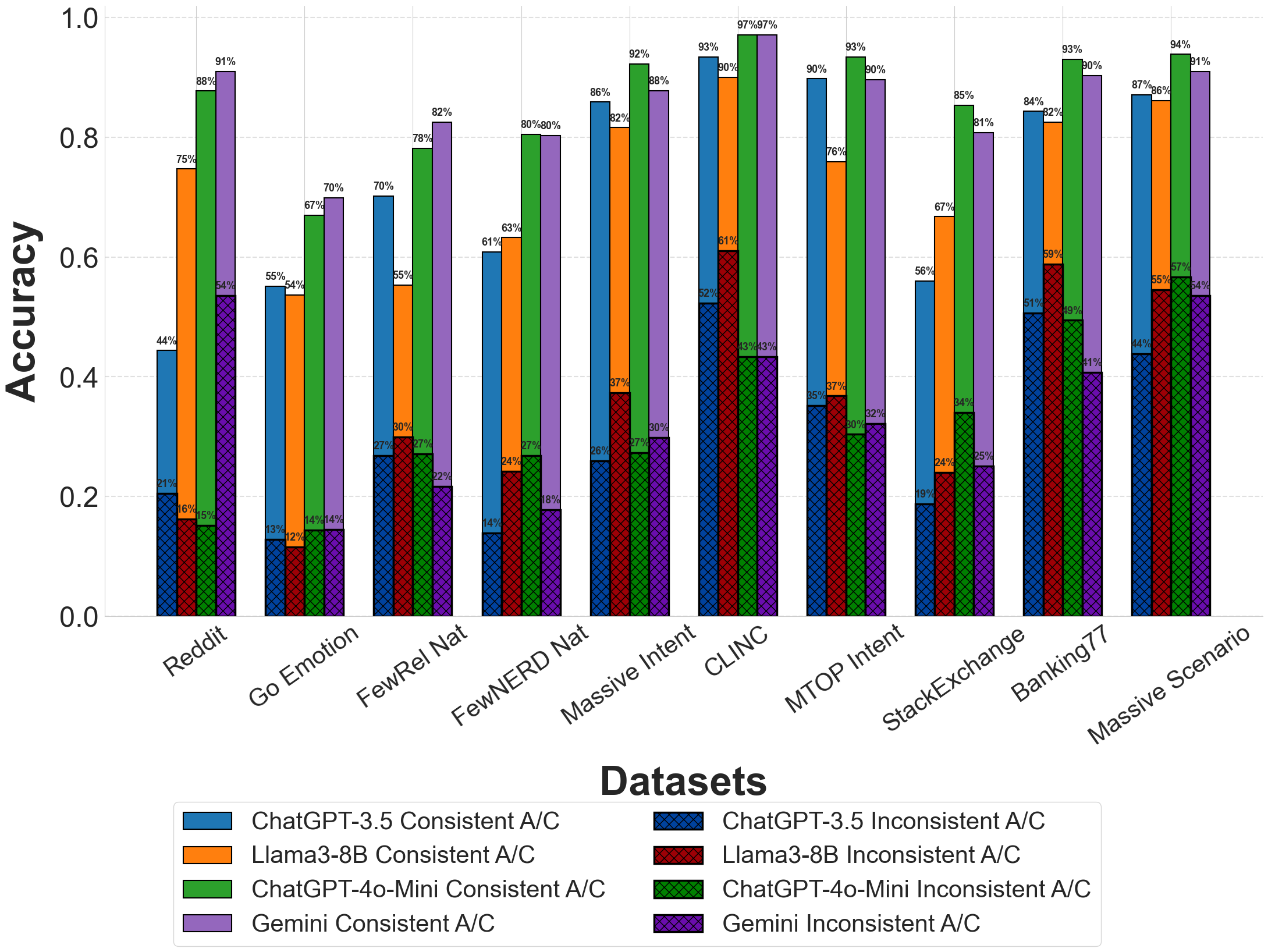}
        \captionsetup{font=footnotesize} 
        \vskip 0.2in
        \caption{\textbf{An illustrative figure highlighting the importance of \textit{consistent-and-inconsistent} sample identification in evaluating LLM performance.} LLM annotations on inconsistent samples (\textbf{dark-colored bars}) exhibit significantly lower accuracy compared to those on consistent samples (\textbf{light-colored bars}). 
        }
        \label{fig:comparison1}

    \end{minipage}
     \hfill
    \label{fig:comparison}
    \begin{minipage}[c]{0.47\textwidth}
        \begin{minipage}[t]{0.4\linewidth}
            \centering
            \includegraphics[width=\linewidth]{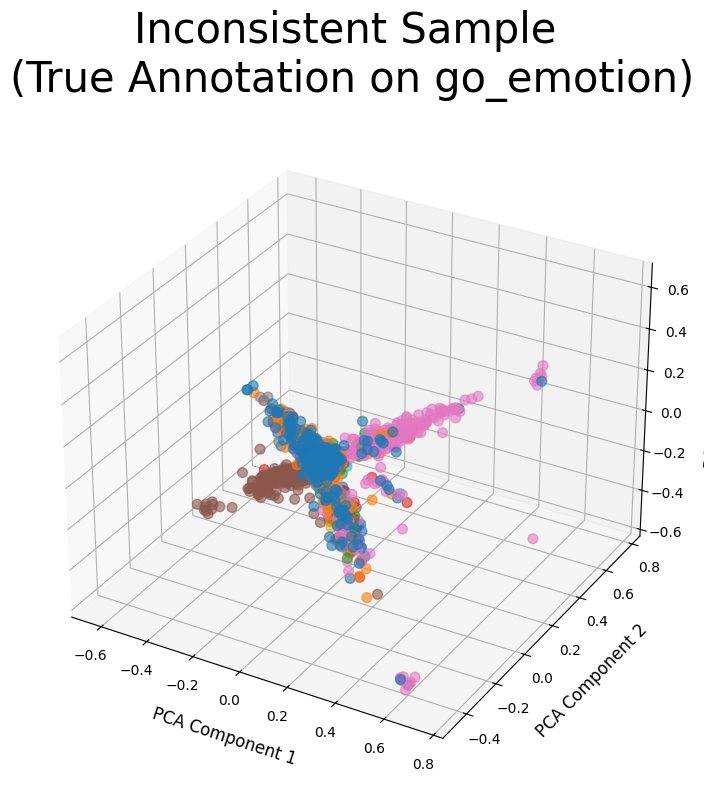}
        \end{minipage}%
        \hskip 0.1in
        \begin{minipage}[t]{0.4\linewidth}
            \centering
            \includegraphics[width=\linewidth]{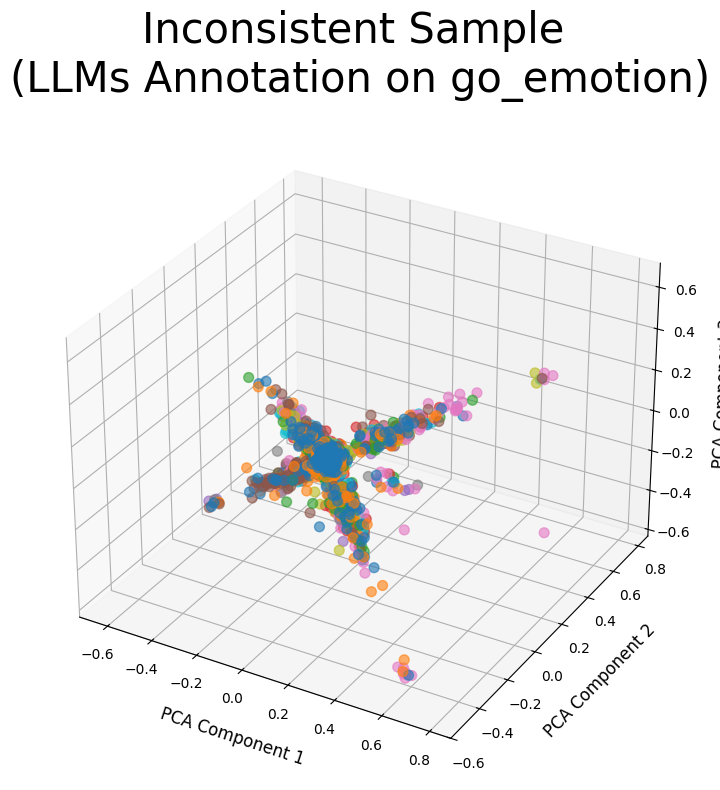}
        \end{minipage}
        
        \begin{minipage}[t]{0.4\linewidth}
            \centering
            \includegraphics[width=\linewidth]{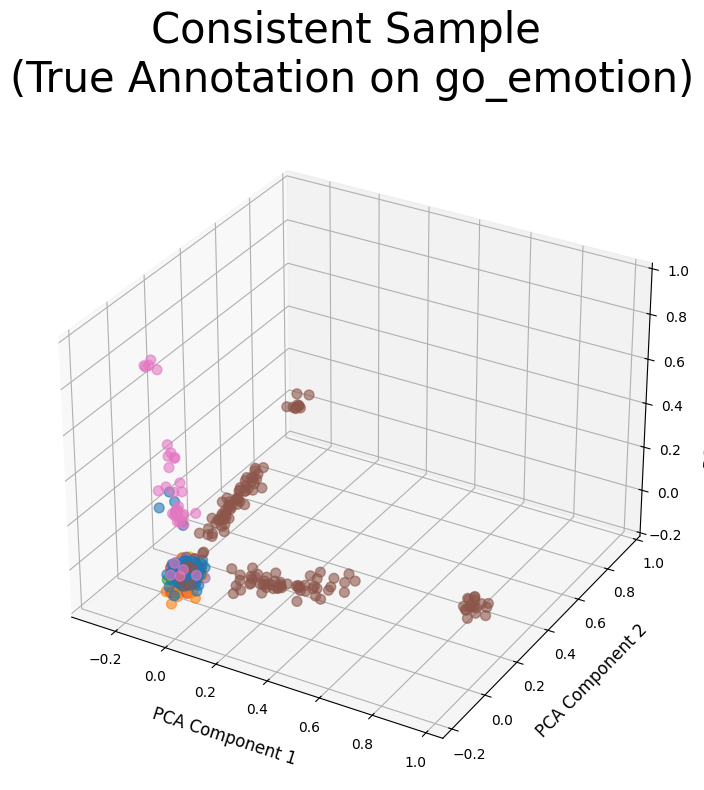}
        \end{minipage}%
        \hskip 0.1in
        \begin{minipage}[t]{0.4\linewidth}
            \centering
            \includegraphics[width=\linewidth]{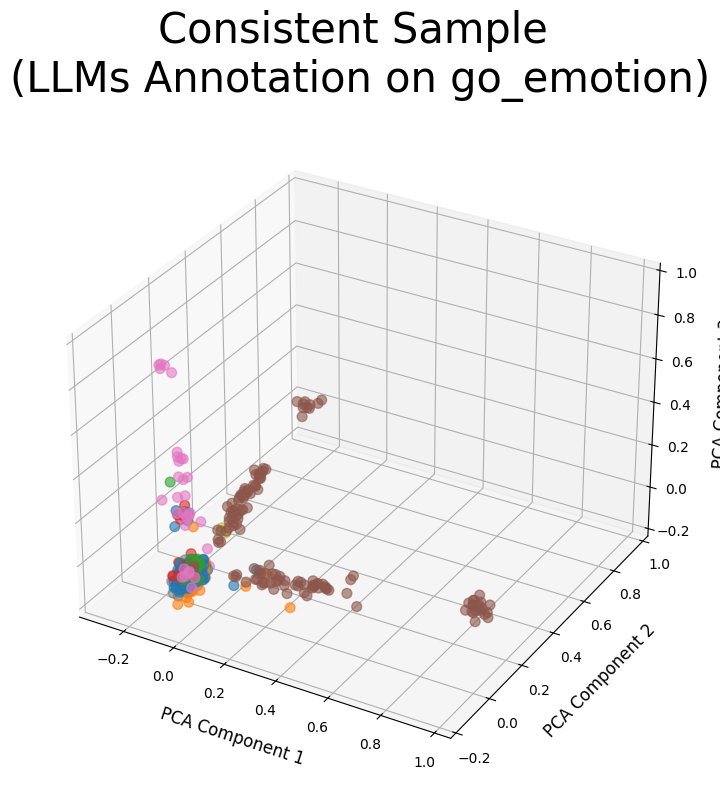}
        \end{minipage}
        \captionsetup{font=footnotesize} 
         \vspace{-9pt}
        \caption{\textbf{Visualization of t-SNE Clustering} (better viewed in color, enlarged) comparing LLM  vs Ground-Truth Annotations on \textit{Go\_Emotion} Dataset. LLM outputs exhibit \textbf{\textit{high similarity}} with ground-truth labels on \textbf{consistent} samples, while showing \textbf{\textit{significant divergence}} on \textbf{inconsistent} samples.
        }
        \label{fig:tsne_clustering}
    \end{minipage}%
    \vskip -0.15in
\end{figure}
\section{Experiments}
\vspace{-0.1in}
\paragraph{Experimental Setup} We collected the CAI Ratio for LLMs—GPT-3.5 Turbo, GPT-4o Mini, Google Gemini 1.5 Flash, and Llama-8B Instruct and evaluated these across ten textual datasets. These datasets include Bank77 \citep{casanueva2020efficient}, CLINC, Go Emotion, MTOP, Massive (Intent) \citep{larson2019evaluation, fitzgerald2022massive, li2020mtop}, StackExchange, Reddit \citep{geigle2021tweac}, FewRel Nat, and FewNerd Nat \citep{han2018fewrel}. Covering domains such as intent classification, topic modeling, and unsupervised intent discovery \citep{zhang2021discovering, zhang2022new}, their annotation practices follow \citep{zhang2023clusterllm}.

\vspace{-0.15in}
\paragraph{Proof-of-Concept Experiments on \textit{Consistency and Inconsistency Identification}}
We first investigate the impact of identifying consistent and inconsistent samples in our framework. Figure ~\ref{fig:comparison1} shows that LLMs achieve significantly higher accuracy on  \textit{consistent} samples, reflecting greater confidence in their predictions, whether observed with a student model or within the LLM's own outputs. The t-SNE visualization in Figure~\ref{fig:tsne_clustering} further confirms that LLM annotations  align closely with ground-truth labels for \textit{consistent samples}, while diverging significantly  for \textit{inconsistent} samples. This contrast highlights the importance of our  identification process for evaluating LLM annotations.
\vspace{-0.15in}
\paragraph{Correlation Results between CAI Ratio and LLM Accuracy} We performed a Pearson correlation analysis to investigate the relationship between CAI Ratio and LLM accuracy. The correlation analysis between the Consistent-over-Inconsistent (CAI) ratio and accuracy  across different LLMs demonstrates a strong relationship between these two metrics. GPT-3.5 shows the highest correlation  ($\rho = 0.93$, $ p = 8.22 \times 10^{-5} $), indicating a very strong positive relationship between CAI and accuracy, with high statistical significance. GPT-4o Mini shows a strong correlation ($\rho = 0.86$, $ p = 1.61 \times 10^{-3} $), suggesting that CAI is a reliable predictor of accuracy for this model. Llama-8B-Instruct  ($\rho = 0.81$, $ p = 1.44 \times 10^{-2} $) and  Google Gemini  ($\rho = 0.72$, $ p = 1.80 \times 10^{-2} $) exhibit  moderate-to-strong correlations with significant statistical confidence. \label{sec:exp_CAI_ratio}
\vspace{-0.15in}
\paragraph{Model Selection with CAI Ratio} Model selection based solely on the CAI Ratio correctly identifies the best-performing LLMs in 60\% of cases. Among the mismatched cases, the accuracy differences are not significant. Although CAI Ratio alone is not a perfect indicator of LLMs accuracy, it serves as a reliable heuristic for selecting well-performing LLMs in unsupervised settings. We have chosen the Best CAI Model and the Best Accuracy Model from the candidate LLM set, which includes GPT-3.5 Turbo, GPT-4o Mini, Google Gemini 1.5 Flash, and Llama-8B Instruct.
\vspace{-0.2in}

\begin{table}[t]
\vspace{-10pt} 
\centering
\scriptsize
\resizebox{0.9\textwidth}{!}{
\begin{tabular}{|l|c|c|c|c|c|c|c|}
\toprule
\multirow{2}{*}{\textbf{Dataset}} & \multicolumn{2}{c|}{\textbf{Best CAI Model}} & \multicolumn{2}{c|}{\textbf{Best Accuracy Model}} & \textbf{Match} & \textbf{Accuracy Difference} \\
\cline{2-5}
& \textbf{Model} & \textbf{Accuracy (\%)} & \textbf{Model} & \textbf{Accuracy (\%)} & & \textbf{(\%)} \\
\midrule
CLINC            & Google Gemini     & 87.24 & Google Gemini     & 87.24 & \ding{51} & 0.00 \\  
MTOP Intent      & Google Gemini     & 75.85 & Google Gemini     & 75.85 & \ding{51} & 0.00 \\  
StackExchange    & Google Gemini     & 57.31 & Google Gemini     & 57.31 & \ding{51} & 0.00 \\  
Banking77        & Google Gemini     & 73.76 & GPT-3.5           & 73.93 & \ding{55} & \textbf{-0.17} \\  
Massive Scenario & Google Gemini     & 67.72 & GPT-3.5           & 75.55 & \ding{55} & \textbf{-7.83} \\  
Reddit           & Google Gemini     & 56.23 & ChatGPT-4o Mini   & 57.39 & \ding{55} & \textbf{-1.16} \\  
Go Emotion       & Google Gemini     & 29.44 & ChatGPT-4o Mini   & 33.82 & \ding{55} & \textbf{-4.38} \\  
FewRel Nat       & Google Gemini     & 52.74 & Google Gemini     & 52.74 & \ding{51} & 0.00 \\  
FewNERD Nat      & Google Gemini     & 75.48 & Google Gemini     & 75.48 & \ding{51} & 0.00 \\  
Massive Intent   & Google Gemini     & 77.03 & Google Gemini     & 77.03 & \ding{51} & 0.00 \\  
\bottomrule
\end{tabular}
}
\vspace{-0.1in} 
\caption{\small \textbf{Model Selection Using CAI Ratio as a Metric}: The model selected based on CAI ratio exhibits a strong correlation with the model achieving the highest accuracy.}
\label{tab:model_selection}
\vskip -0.25in
\end{table}
\section{Conclusion}
\vspace{-0.1in}
In this work, we propose a novel and effective metric, the \textbf{CAI Ratio}, based on a {Agentic Annotation Evaluation Paradigm} for {unsupervised dataset annotation} aligned with user preferences. The CAI Ratio has demonstrated its effectiveness in both \textbf{LLM annotation evaluation} and \textbf{model selection}. Evaluated on ten domain-specific NLP datasets, the CAI metric exhibited a strong positive correlation with LLM performance, confirming its efficacy as a {model selection and evaluation tool} for unsupervised dataset annotation tailored to user preferences.
\newpage
\bibliography{iclr2025_conference}
\bibliographystyle{iclr2025_conference}
\newpage
\appendix
\section{Appendix}
This supplementary material is organized as follows. In Sec~\ref{appendix:CAI}, we provide a detailed interpretation of the CAI Ratio, highlighting its advantages and distinctions compared to traditional evaluation metrics. In Sec~\ref{appendix:experiment}, we present comprehensive experimental results, including model accuracy and CAI ratios across various datasets evaluated with different LLMs. Finally, in Sec~\ref{appendix:tsne_visualization}, we showcase t-SNE visualization results, illustrating clustering patterns for consistent and inconsistent samples on additional datasets.

\label{appendix:CAI}
\subsection{\textbf{Consistent and Inconsistent Ratio Interpretation}}
The CAI ratio provides a principled means to assess the reliability of LLM-generated annotations in the absence of labelled supervision. A significantly higher CAI ratio (\( \text{CAI Ratio} \gg 1 \)) may indicate consistency or higher annotation accuracy, while a lower CAI ratio (\( \text{CAI Ratio} \ll 1 \)) suggests greater lower annotation accuracy and inconsistency in the LLM’s outputs.  
In these cases, 
the ratio suggests that the LLM's outputs are unreliable, necessitating refinements with external human annotations or additional prior knowledge to improve annotation accuracy. 

Furthermore, the relationship between the CAI ratio and LLM annotation accuracy can be formalized as the \textit{Law of Consistency}. This principle states that if both the LLM and student model are optimal hypotheses, denoted as $ T^{*} $ and $ S^{*} $ for a given dataset $ D_u $, the number of consistent samples should asymptotically exceed the number of inconsistent samples as the dataset size approaches infinity.
[Law of Consistency]
Let $ T^* $ and $ S^* $ be the optimal teacher (LLM) and student model hypotheses for an unsupervised dataset $ D_u $. Define $ N_C $ and $ N_{IC} $ as the number of consistent and inconsistent samples, respectively, identified by the CAI ratio. As the dataset size $ |D_u| \to \infty $, the probability that that of consistent samples surpasses the number of inconsistent samples approaches one:
    $\lim_{|D_u| \to \infty} P(N_C > N_{IC}) = 1$.

\subsection{Comparison with Traditional Evaluation Metrics}
\begin{table}[h]
    \centering
    \small 
    \renewcommand{\arraystretch}{1.1} 
    \setlength{\tabcolsep}{4pt} 
    \begin{tabular}{lccc}
        \toprule
        \textbf{Metric} & \textbf{Ground-Truth Labels?} & \textbf{Data Drift?} & \textbf{Tracks Annotation Quality Over Time?} \\
        \midrule
        Accuracy & \cmark  & \xmark  & \xmark  \\
        Precision/Recall & \cmark  & \xmark  & \xmark  \\
        F1-score & \cmark  & \xmark  & \xmark  \\
        CAI Ratio & \xmark  & \cmark  & \cmark  \\
        \bottomrule
    \end{tabular}
    \caption{Comparison of Traditional Metrics and CAI Ratio}
    \label{tab:metrics_comparison}
\end{table}
\subsection{\textbf{Pearson Correlation Test for Consistent and Inconsistent Ratio}}
We have performed a Pearson correlation, the correlation coefficient $ r$ is calculated as:
\[
r = \frac{\sum_{i=1}^n (x_i - \bar{x})(y_i - \bar{y})}{\sqrt{\sum_{i=1}^n (x_i - \bar{x})^2} \sqrt{\sum_{i=1}^n (y_i - \bar{y})^2}}
\]
where $ x_i$ symbolises the CAI Ratio. $ y_i$ denotes the LLM annotation accuracies.
$\bar{x}$ and $ \bar{y}$ are the average mean of $ x_i$ and $ y_i$, accordingly. $ n$ is the number of samples we have used for evaluation. To assess the statistical significance, we use a hypothesis test for the correlation coefficient, calculating a t-statistic \citep{schober2018correlation}:
\[
t = r \sqrt{\frac{n - 2}{1 - r^2}}
\]
The P-value is then calculated from the t-distribution with $ n - 2$ degrees of freedom.
\begin{tabular}{|l|c|c|}
\hline
\centering
\textbf{LLM} & \textbf{Pearson Correlation ($\rho$)} & \textbf{p-value ($p$)} \\
\hline
GPT-3.5 & 0.93 & $8.22 \times 10^{-5}$ \\
GPT-4o Mini & 0.86 & $1.61 \times 10^{-3}$ \\
Llama-8B-Instruct & 0.81 & $1.44 \times 10^{-2}$ \\
Google Gemini & 0.72 & $1.80 \times 10^{-2}$ \\
\hline
\end{tabular}
\subsection{\textbf{Importance of Student Model in Agentic Annotation Evaluation Paradigm}}
The inclusion of the student model is essential as it provides a safeguard against underperformance by the LLM. Additionally, the student model serves as a reference point for "course tracking," meaning that it allows us to monitor and guide the annotation process by comparing the student model's output with the teacher model's output. This is particularly evident in our experiments where the Meta-8B Instruct model, serving as a low-competency "noisy teacher," exhibited suboptimal performance on most of the eight datasets, as reflected in its low CAI scores. The student model addresses this challenge by collaborating with the teacher model to iteratively refine annotations, ensuring robustness even when the teacher model lacks competency in specific tasks. We justify the necessity of the student model through experimental analysis.Another key motivation for our approach, including the use of a student model, is to enhance efficiency. This efficiency is evident in two significant aspects:
\begin{itemize}
    \item Computational Efficiency: Our method requires access to the teacher model only twice per dataset, with minimal or no reliance on demonstrations for prompting. This substantially reduces computational overhead.
    \item Cost-Effectiveness: For closed-source models with API service fees, our approach offers a cost-efficient solution. By utilizing the student model alongside our proposed clustering operation and a limited number of teacher model predictions, our method achieves superior performance compared to both models individually. Importantly, it does so at a lower cost, particularly when compared to methods that rely on iterative self-correction.
\end{itemize}

\section{\textbf{Experimental Results}}
\label{appendix:experiment}
\paragraph{Implementation Details}
The top-k selection and proportions of consistent and user-preference samples are as follows. For CLINC and Massive Scenario, `top-k` is set to 5, with `proportion` at 0.2. For MTOP Intent, `proportion` is set to 1, and `top-k` is updated to 15 after printing the current value. In StackExchange, `top-k` is set to 5 and `proportion` to 1, while in Banking77, `top-k` is set to 3 and `proportion` is 0.2. In massive intent, `top-k` is 20 and `proportion` is 0.5), proportion=0.2, and few real nat has top-k=30, and proportion is 1. In 'reddit', `top-k` is set to 7, and the proportion is 0.2. All tests are done with two random seeds with temperature parameters (0.5 and 1) for user preference samples, student model-assigned annotation, and LLMs with and without student annotations.

\begin{table}[H]
\centering
\renewcommand{\arraystretch}{1.1}
\resizebox{1\textwidth}{!}{%
\begin{tabular}{|l|c|c|c|c|c|c|c|c|}
\hline
\small
\multirow{2}{*}{\textbf{Dataset}} & \multicolumn{2}{c|}{\textbf{GPT-3.5}} & \multicolumn{2}{c|}{\textbf{ChatGPT-4o Mini}} & \multicolumn{2}{c|}{\textbf{Google Gemini}} & \multicolumn{2}{c|}{\textbf{Llama-8B}} \\
\cline{2-9}
 & \textbf{Accuracy (\%) $\pm$ Std} & \textbf{CAI Ratio} & \textbf{Accuracy (\%) $\pm$ Std} & \textbf{CAI Ratio} & \textbf{Accuracy (\%) $\pm$ Std} & \textbf{CAI Ratio} & \textbf{Accuracy (\%) $\pm$ Std} & \textbf{CAI Ratio} \\
\hline
Banking77        & 73.93 $\pm$ 0.81  & 1.46  & 65.78 $\pm$ 0.24  & 1.35  & 73.73 $\pm$ 0.03  & 5.34  & 33.06 $\pm$ 1.92  & 0.68  \\
Clinc            & 79.01 $\pm$ 1.08  & 1.55  & 81.46 $\pm$ 0.36  & 1.99  & 87.50 $\pm$ 0.26  & 10.90  & 32.49 $\pm$ 6.73  & 0.56  \\
Massive Scenario & 75.55 $\pm$ 1.76  & 1.39  & 66.83 $\pm$ 1.31  & 1.23  & 67.95 $\pm$ 0.23  & 3.41  & 43.52 $\pm$ 1.85  & 0.67  \\
MTOP Intent      & 52.49 $\pm$ 2.52  & 0.68  & 74.54 $\pm$ 0.32  & 0.72  & 75.61 $\pm$ 0.23  & 2.94  & 34.17 $\pm$ 6.70  & 0.35  \\
Stack Exchange   & 32.27 $\pm$ 0.65  & 0.40  & 51.90 $\pm$ 0.18  & 0.30  & 57.48 $\pm$ 0.17  & 2.11  & 11.02 $\pm$ 2.78  & 0.23  \\
Reddit           & 51.12 $\pm$ 1.27  & 0.50  & 57.39 $\pm$ 0.40  & 0.41  & 56.73 $\pm$ 0.50  & 3.10  & 36.31 $\pm$ 0.97  & 0.333  \\
Go Emotion       & 31.84 $\pm$ 0.87  & 0.12  & 33.82 $\pm$ 0.25  & 0.12  & 29.72 $\pm$ 0.28  & 0.81  & 22.53 $\pm$ 0.21  & 0.102  \\
Few Rel Nat      & 32.87 $\pm$ 1.72  & 0.28  & 35.87 $\pm$ 0.22  & 0.26  & 52.96 $\pm$ 0.21  & 1.70  & 14.25 $\pm$ 0.36  & 0.128  \\
Few Nerd Nat     & 47.70 $\pm$ 1.36  & 0.42  & 62.20 $\pm$ 0.19  & 0.30  & 75.35 $\pm$ 0.13  & 2.37  & 17.60 $\pm$ 2.02  & 0.055  \\
Massive Intent   & 71.52 $\pm$ 0.95  & 1.62  & 76.93 $\pm$ 0.16  & 1.47  & 76.90 $\pm$ 0.13  & 5.41  & 45.41 $\pm$ 0.06  & 0.730  \\
\hline
\end{tabular}
}
\end{table}
\begin{table}[H]
\centering
\caption{Model Selection Results Using CAI as a Metric}
\label{tab:model_selection_}
\resizebox{1\columnwidth}{!}{
\begin{tabular}{|l|c|c|c|c|c|c|c|c|c|c|c|}
\toprule
         \textbf{Dataset} &  \textbf{GPT-3.5 CAI} &  \textbf{ChatGPT-4o Mini CAI} &  \textbf{Google Gemini CAI} &  \textbf{Llama 8B CAI} &  \textbf{GPT-3.5 Accuracy} &  \textbf{ChatGPT-4o Mini Accuracy} &  \textbf{Google Gemini Accuracy} &  \textbf{Llama 8B Accuracy} & \textbf{Best CAI Model} & \textbf{Best Accuracy Model} \\
\midrule
           CLINC &         1.55 &               1.9974 &             10.900 &         0.560 &             79.01 &                     81.46 &                   87.24 &              32.49 &  Google Gemini &       Google Gemini \\
     MTOP Intent &         0.68 &               0.7236 &              2.940 &         0.670 &             52.49 &                     74.54 &                   75.85 &              43.52 &  Google Gemini &       Google Gemini \\
   StackExchange &         0.40 &               0.3014 &              2.110 &         0.350 &             32.27 &                     51.90 &                   57.31 &              34.17 &  Google Gemini &       Google Gemini \\
       Banking77 &         1.46 &               1.3494 &              3.545 &         0.680 &             73.93 &                     65.78 &                   73.76 &              33.06 &  Google Gemini &             GPT-3.5 \\
Massive Scenario &         1.39 &               1.2269 &              4.375 &         0.230 &             75.55 &                     66.83 &                   67.72 &              11.02 &  Google Gemini &             GPT-3.5 \\
          Reddit &         0.50 &               0.4151 &              3.100 &         0.333 &             51.12 &                     57.39 &                   56.23 &              36.31 &  Google Gemini &     ChatGPT-4o Mini \\
      Go Emotion &         0.12 &               0.1238 &              0.810 &         0.102 &             31.84 &                     33.82 &                   29.44 &              22.53 &  Google Gemini &     ChatGPT-4o Mini \\
      FewRel Nat &         0.28 &               0.2613 &              1.700 &         0.128 &             32.87 &                     35.87 &                   52.74 &              14.25 &  Google Gemini &       Google Gemini \\
     FewNERD Nat &         0.42 &               0.3064 &              2.370 &         0.055 &             47.70 &                     62.20 &                   75.48 &              17.60 &  Google Gemini &       Google Gemini \\
  Massive Intent &         1.62 &               1.4701 &              5.410 &         0.730 &             71.52 &                     76.93 &                   77.03 &              45.41 &  Google Gemini &       Google Gemini \\
\bottomrule
\end{tabular}
}
\end{table}

\begin{table}[H]
\centering
\caption{Accuracy with Consistent Samples and Inconsistent Samples Across Four LLMs}
\label{tab:accuracy_consistent_inconsistent}
\resizebox{\textwidth}{!}{  
\begin{tabular}{lcccccccc}
\toprule
\textbf{Dataset} & \multicolumn{2}{c}{\textbf{ChatGPT-3.5}} & \multicolumn{2}{c}{\textbf{Llama-8B}} & \multicolumn{2}{c}{\textbf{ChatGPT-4o Mini}} & \multicolumn{2}{c}{\textbf{Google Gemini}} \\
\cmidrule(lr){2-3} \cmidrule(lr){4-5} \cmidrule(lr){6-7} \cmidrule(lr){8-9}
 & Consistent (\%) & Inconsistent (\%) & Consistent (\%) & Inconsistent (\%) & Consistent (\%) & Inconsistent (\%) & Consistent (\%) & Inconsistent (\%) \\
\midrule
Reddit         & 44.37 & 20.53 & 74.68 & 16.21 & 87.70 & 15.15 & 86.37 & 13.66 \\
Go Emotion     & 55.10 & 12.87 & 53.61 & 11.57 & 66.93 & 14.38 & 69.88 & 14.48 \\
FewRel Nat     & 70.16 & 26.86 & 55.32 & 29.97 & 78.11 & 27.11 & 82.47 & 21.67 \\
FewNERD Nat    & 60.82 & 13.94 & 63.21 & 24.17 & 80.48 & 26.83 & 80.27 & 17.82 \\
Massive Intent & 85.86 & 25.92 & 81.59 & 37.34 & 92.26 & 27.30 & 87.79 & 29.81 \\
CLINC          & 93.37 & 52.27 & 90.02 & 61.04 & 97.09 & 43.33 & 90.98 & 53.53 \\
MTOP Intent    & 89.75 & 35.21 & 75.88 & 36.78 & 93.42 & 30.38 & 89.63 & 32.19 \\
StackExchange  & 55.99 & 18.76 & 66.71 & 24.01 & 85.32 & 34.05 & 80.77 & 25.07 \\
Banking77      & 84.30 & 50.64 & 82.48 & 58.77 & 93.01 & 49.43 & 90.31 & 40.77 \\
Massive Scenario & 87.04 & 43.86 & 86.09 & 54.55 & 93.83 & 56.65 & 90.98 & 53.54 \\
\bottomrule
\end{tabular}}
\end{table}
\end{document}